\documentclass[10pt,letterpaper]{article}

\usepackage{ccn}
\usepackage{pslatex}
\usepackage{apacite}
\usepackage{xcolor}
\usepackage{graphicx}

\title{One system for learning and remembering episodes and rules}

\author{{\large \bf Joshua T. S. Hewson (joshua\_hewson@brown.edu)} \\
  Department of Cognitive, Linguistic, and Psychological Sciences, Brown University,
Providence, RI
  \AND {\large \bf Sabina J. Sloman (sabina.sloman@manchester.ac.uk)} \\
  Department of Computer Science, University of Manchester, Manchester, UK
\AND {\large \bf Marina Dubova (mdubova@iu.edu)} \\
  Cognitive Science Program, Indiana University, Bloomington, IN}

\begin{document}

\begin{titlepage}
    \maketitle
\end{titlepage}

\section{Abstract}
{
\bf
Humans can learn individual episodes and generalizable rules and also successfully retain both kinds of acquired knowledge over time. In the cognitive science literature, (1) learning individual episodes and rules and (2) learning and remembering are often both conceptualized as competing processes that necessitate separate, complementary learning systems.
Inspired by recent research in statistical learning, we challenge these trade-offs, hypothesizing that they arise from capacity limitations rather than from the inherent incompatibility of the underlying cognitive processes. Using an associative learning task, we show that one system with excess representational capacity can learn and remember both episodes and rules.
}
\begin{quote}
\small
\textbf{Keywords: Remembering, catastrophic forgetting, complementary learning systems, continual learning} 

\end{quote}

\section{Introduction}
% 5. We also uncover interesting details about the dynamics of forgetting, etc -- important for studying further in the context of continual learning

In the study of learning, two trade-offs have historically been observed in the behavior of computational models: (1) between the abilities to simultaneously learn individual episodes and generalizable rules and (2) between the abilities to learn and to remember.
For example, connectionist models often show that (1) memorizing individual episodes leads to a reduced ability to learn the rules required to generalize to new episodes (``overfitting") \cite{mcclelland1995complementary} and (2) learning in a new task leads to catastrophic forgetting of what has been learned in previous tasks \cite{mccloskey1989catastrophic}.
These observations motivated the creation of dual-system theories, such as the complementary learning systems model \cite{mcclelland1995complementary}, which posit separate learning systems for learning and remembering episodes and rules.

Recent research has shown that the trade-off between learning episodes and rules is not inherent to learning in computational systems.
The computational models in which these trade-offs were historically observed had limited \textit{capacity}: They could memorize only a small number of their observations.
Computational systems with excess capacity -- which can recover far more relationships between the features of observations -- have the ability to both memorize and generalize, i.e., to learn both episodes and rules \cite{dubova2023excess, belkin2019reconciling, nakkiran2019deep, davies2023unifying}.
In this study, we demonstrate that  excess capacity systems can also overcome the apparent trade-off between learning and remembering, i.e., they can simultaneously successfully learn new episodes and rules \textit{and} remember previously-learned episodes and rules.
% ...generalize, memorize, and retain their memories.

\section{Methods}
   \paragraph{Catastrophic forgetting.}
        Human participants in the behavioral test referenced by \citeA{mcclelland1995complementary} were tasked with memorizing batches of random word pairings in a blocked regime \cite{barnes1959fate}.
        During the first block, participants were presented with a list of words (list $A$) and tasked with memorizing arbitrary associations between the words on list $A$ and the words on another list $B$ ($A-B$ pairings).
        During the second block, they were presented with a new word list $C$ and tasked with memorizing arbitrary associations between the words on list $A$ and on list $C$ ($A-C$ pairings).
        Over the course of training on the $A-C$ pairings, participants were tested on the $A-B$ pairings they learned during the first block. Participants showed memory interference, but were still able to retain most of the previously learned associations. \citeA{mccloskey1989catastrophic} modeled behavior in this task with a simple connectionist model. 
        This model forgot nearly all information about the $A-B$ pairings after being trained on the $A-C$ pairings, a phenomenon they referred to as \textit{catastrophic forgetting}.

\paragraph{Model. }Inspired by \citeA{mccloskey1989catastrophic}, we used a simple multi-layer perceptron architecture with two hidden layers of equal width.\footnote{Find our code at: \url{https://github.com/TheLemonPig/ECLvsCLS}}

\paragraph{Task. }
We expand on \citeA{mccloskey1989catastrophic}'s procedure by changing the data to vary on a continuum from rules to episodes, so that the dynamics of learning and forgetting of arbitrary associations between episodes and generalizable rules can be studied together.

Two sample datasets of 10 5-dimensional samples, $A_{train}$ and $A_{test}$, were created by sampling from a Gaussian probability distribution. These datasets were then passed through a transformation $f$.
Two target datasets, $B$ and $C$, were each formed by taking a weighted sum between the transformed sample data and a set of random perturbations. A third target dataset $D$ was created by removing the random perturbations from $C$:
\begin{center}
    $A_{train} \sim \mathcal{N}(0,1)$\\ $A_{test} \sim \mathcal{N}(0,1)$\\
    $B = (1 - noise) \cdot f(A_{train}) + noise \cdot \epsilon_B$\\
    $C = (1 - noise) \cdot f(A_{test}) + noise \cdot \epsilon_C$\\
    $D = (1 - noise) \cdot f(A_{test})$\\
\end{center}
where $0\le noise\le 1$, $\epsilon_B \sim \mathcal{N}(0, 1)$ and $\epsilon_C \sim \mathcal{N}(0, 1)$.
$f$ represents the generalizable rule that characterizes the relationship between the sample and corresponding target data (i.e., that characterizes each of the $A_{train} - B$, $A_{test} - C$ and $A_{test} - D$ pairings). 
The $noise$ parameter controls the amount of structure in the data: When $noise = 0$, the task amounts entirely to learning of the generalizable rule; when $noise = 1$, the task amounts entirely to learning arbitrary associations between the sample data and episodes $\epsilon_B$ (in the $A_{train} - B$ pairings) and $\epsilon_C$ (in the $A_{test} - C$ pairings).

\paragraph{Capacity. }
Our key manipulation was the \textit{capacity} of each model we tested.
The capacity of a model is defined as the minimum number of hidden nodes needed to fully memorize a given dataset.
\textit{Constrained capacity} models have fewer hidden nodes than required to memorize the data they are presented with.
\textit{Sufficient capacity} models have just enough nodes to memorize the data they are presented with.
\textit{Excess capacity} models have more nodes than required to memorize the data they are presented with.

We tested models with capacities of .5, 1, 10 and 100 times the capacity needed to fully memorize the datasets (constrained, sufficient, excess and excess capacity, respectively).
\paragraph{Training. }
During Block 1, the models were trained to associate $A_{train}$ with $B$, which involves learning both the rule $f$ and the arbitrary component of the episodes, $\epsilon_B$. During Block 1, we also tested the models' abilities to generalize from $A_{test}$ to $C$.
During Block 2, the models were trained to associate $A_{train}$ with $C$. During Block 2, we also tested the models' abilities to recall the $A_{train}-B$ pairings and to predict the $A_{test}-D$ pairings, which capture the models' abilities to remember episodes and rules, respectively (we test learning of the rule on the basis of performance on the $A_{test}-D$ pairings in order to isolate error from failure to learn $f$ from error caused by the noise added to $C$).

The models were optimized with Stochastic Gradient Descent using a mean squared error loss function (learning rate = $0.01$). All models were trained until convergence, defined as a rate of decrease in loss going below $1\times10^{-5}$ per $5,000$ epochs.
We ran all simulations 100 times.
\begin{figure}
    \begin{center}
        \centering
        \small
        \includegraphics[width=0.4\textwidth]{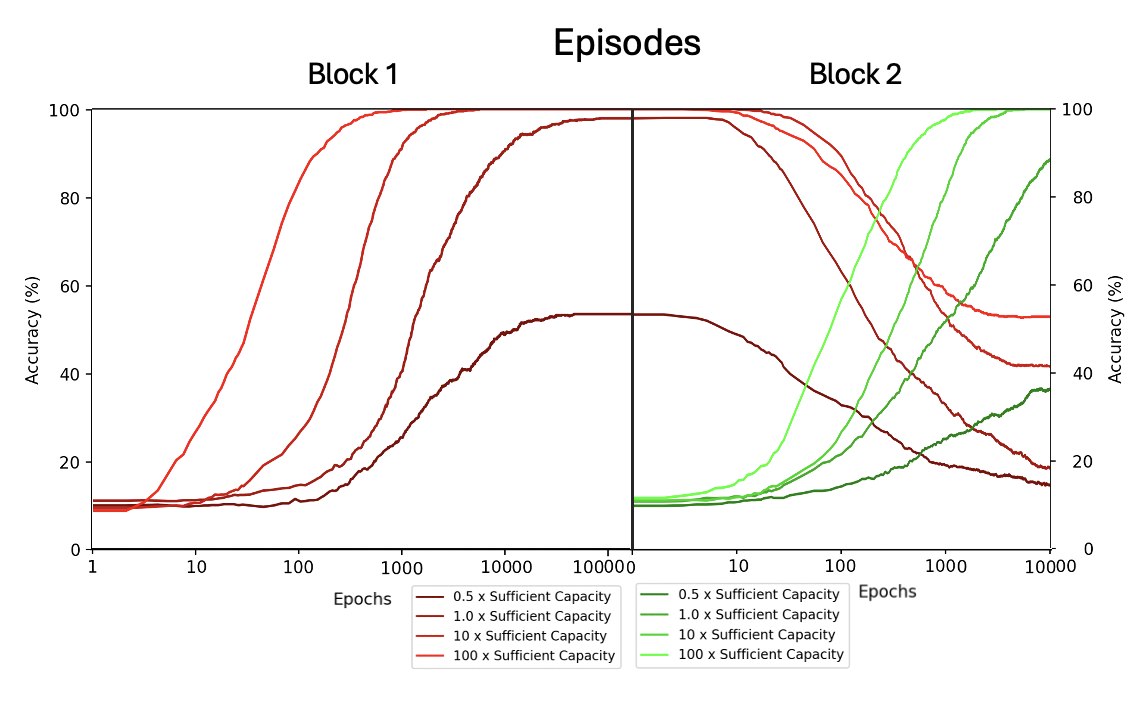}
        \includegraphics[width=0.4\textwidth]{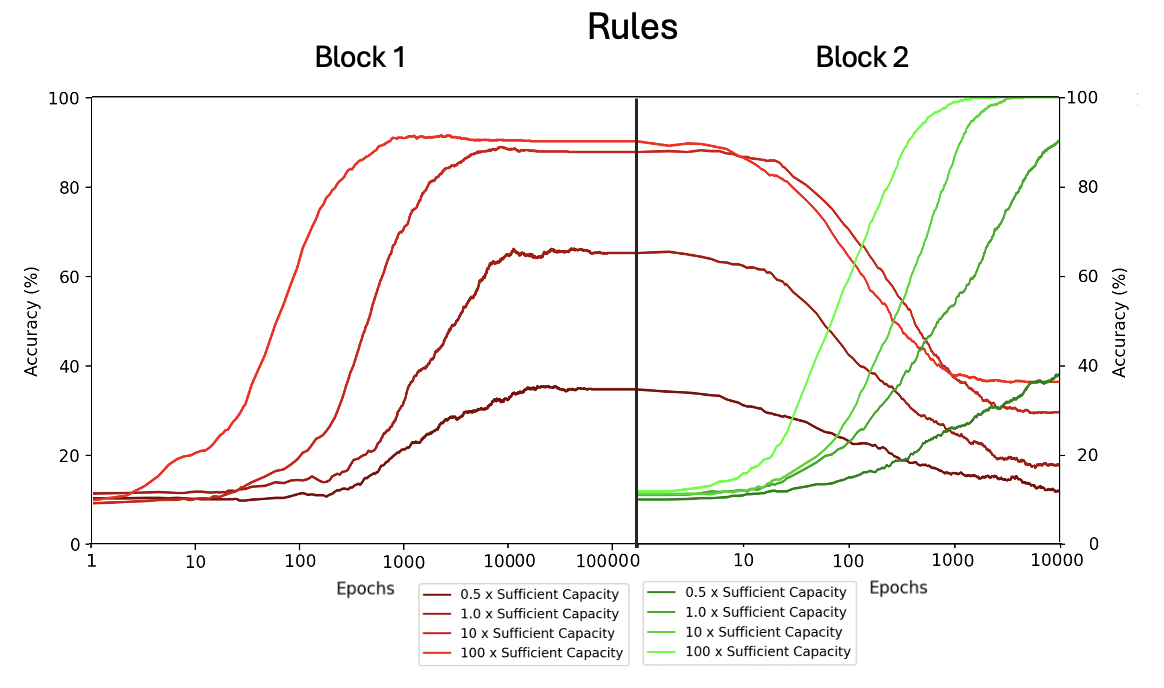}
    \end{center}
    \caption{Temporal plots for mean classification accuracy over training (the noise level is fixed at 25\%). Left: The episode (top) or rule (bottom) for $A_{train}-B$ is learned. Right: The episode $A_{train}-C$ is learned (green lines) while the episode (top) or rule (bottom) for $A_{train}-B$ is being forgotten.} 
\label{fig:image}
\end{figure}

\begin{figure}[t]
    \begin{center}
        \centering
        \small
        \includegraphics[width=0.35\textwidth]{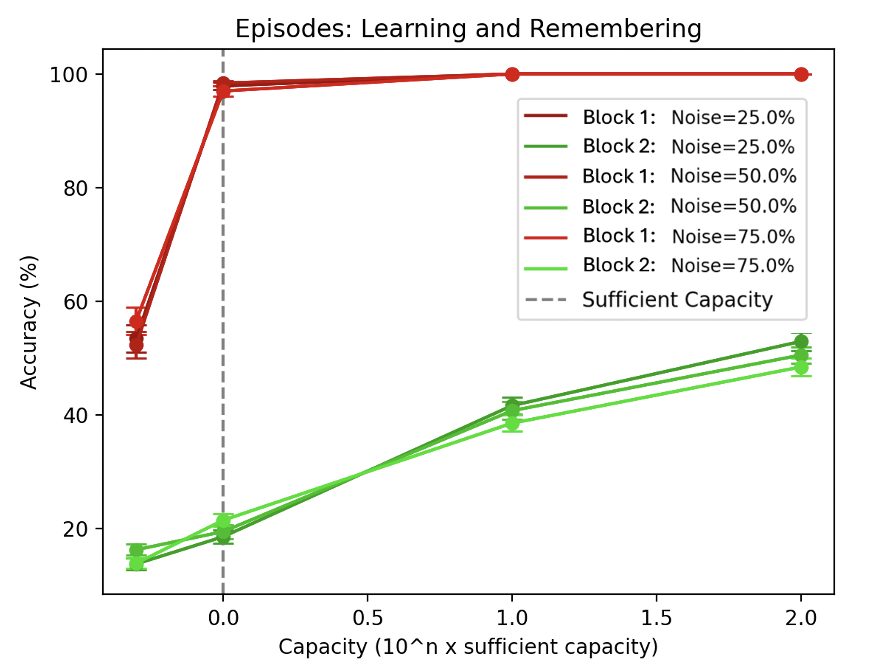}
        \includegraphics[width=0.35\textwidth]{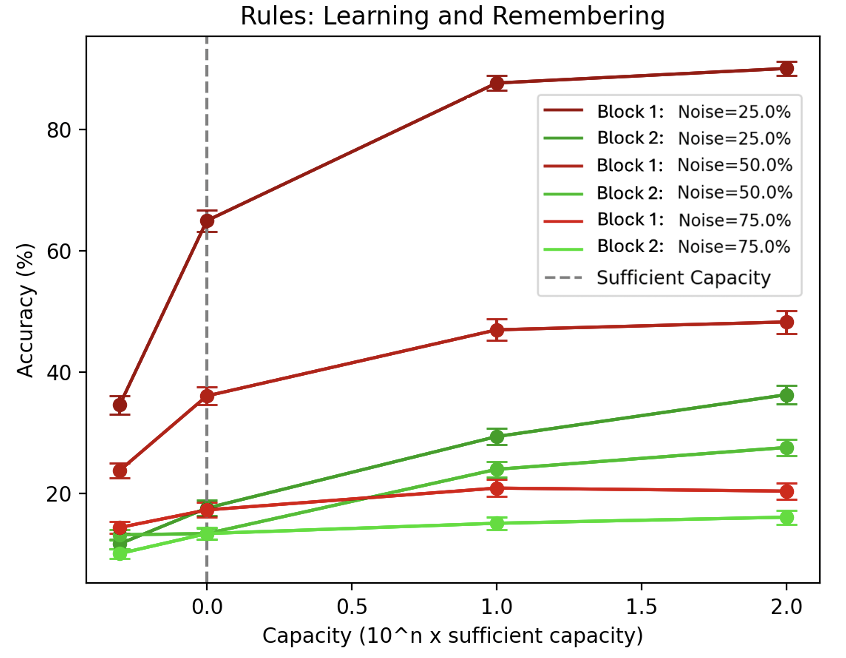}
    \end{center}
    \caption{Final averaged mean results after training on Block 1 and 2 respectively, with varying levels of noise. Left of the dashed line: constrained capacity; Dashed line: sufficient capacity, Right of the dashed line: excess capacity. Error bars show standard errors.} 
\label{fig:image}
\end{figure}
\section{Results}
Consistent with prior literature (e.g., \citeA{belkin2019reconciling, nakkiran2019deep}), in all cases, the models with excess capacity were better able than models with constrained and sufficient capacity to learn both episodes and rules  (Fig. 2). The difference between the mean classification accuracy of all models was statistically significant ($p < 0.001$) for both learning episodes, defined by the t-test of the means between each pair of models, at the end of training.
In other words, there was not a consistent trade-off between learning episodes and learning rules in the excess capacity regime.

Consistent with prior work on catastrophic forgetting, models with constrained and sufficient capacity exhibited only a limited ability to retain prior knowledge when having to learn a new set of interfering associations. The models with excess capacity were better able to retain knowledge of both episodes and rules (Figs. 1 and 2). All pairwise comparisons between the means of performance of excess vs. constrained/sufficient capacity models were significant at the $p<.001$ level.\\
\section{Conclusion}
Our results demonstrate the in-principle ability of one computational learning system to both learn and remember episodes and rules. By challenging the traditional view of learning and remembering episodes and rules as inherently competing processes, this work opens new avenues for understanding the flexibility and nuance of cognitive function by exploring the properties of learning in different capacity regimes.
Our findings also have important implications for the study of continual learning, transfer learning, and the development of more advanced cognitive architectures \cite{mannering2021catastrophic,vandeven2024continual,achille2019critical,sherman2023multiple,schapiro2017complementary,liu2022towards}.

\bibliographystyle{apacite}

\setlength{\bibleftmargin}{.125in}
\setlength{\bibindent}{-\bibleftmargin}

\bibliography{ccn_style}

\end{document}